# Developing a Multilingual Annotated Corpus of Misogyny and Aggression

**Shiladitya Bhattacharya[1], Siddharth Singh[2], Ritesh Kumar[2], Akanksha Bansal[3], Akash Bhagat[2], Yogesh Dawer[2], Bornini Lahiri[4], Atul Kr. Ojha[3]**

[1]Jawaharlal Nehru University, New Delhi, [2]Dr. Bhimrao Ambedkar University, Agra
[3]Panlingua Language Processing LLP, New Delhi, [4]Indian Institute of Technology, Kharagpur
comma.kmi@gmail.com

## Abstract

In this paper, we discuss the development of a multilingual annotated corpus of misogyny and aggression in Indian English, Hindi, and Indian Bangla as part of a project on studying and automatically identifying misogyny and communalism on social media (the ComMA Project). The dataset is collected from comments on YouTube videos and currently contains a total of over 20,000 comments. The comments are annotated at two levels - aggression (overtly aggressive, covertly aggressive, and non-aggressive) and misogyny (gendered and non-gendered). We describe the process of data collection, the tagset used for annotation, and issues and challenges faced during the process of annotation. Finally, we discuss the results of the baseline experiments conducted to develop a classifier for misogyny in the three languages.

## 1. Introduction

The proliferation in Social Networking (platforms and users) has transformed our communities and the manner in which we communicate. One of the widespread impact can be seen through the hate that has been vocalised through platforms like Facebook, Twitter, and YouTube, where content sharing and communication are integrated together. The hate itself is not new but its expression is a matter of concern for the following reasons:

- the articulation is strong enough to break down the community ties;

- the impact of such articulation travels from online to offline domain and has led to incidents like riots, lynching, killing, and ostracisation of individuals and communities;

- the articulation is now manipulated to impact electoral processes, commercial ventures, individual reputation, and communal ties.

Thus, it has become all the more important for scholars and researchers to take the initiative and find methods to identify and compile the source and articulation of aggression and strategise ways to control and curb them. The eventual idea is to structure the online community much like the way law and order is maintained in any geographical community. It is for this reason that we have initiated the building of a sizeable corpus comprising YouTube comments to identify the *modus operandi* behind encoding of misogyny and aggression in user-generated posts and automatically identify those.

In recent times, there has been a large number of studies exploring different aspects of hateful and aggressive language and their computational modelling and automatic detection such as toxic comments[1], trolling (Cambria et al., 2010; Kumar et al., 2014; Mojica de la Vega and Ng, 2018; Mihaylov et al., 2015), flaming / insults (Sax, 2016; Nitin et al., 2012), radicalization (Agarwal and Sureka, 2015; Agarwal and Sureka, 2017), racism (Greevy and Smeaton, 2004; Greevy, 2004), online aggression (Kumar et al., 2018a), cyberbullying (Xu et al., 2012; Dadvar et al., 2013), hate speech (Kwok and Wang, 2013; Djuric et al., 2015; Burnap and Williams, 2015; Davidson et al., 2017; Malmasi and Zampieri, 2017; Malmasi and Zampieri, 2018), abusive language (Waseem et al., 2017; Nobata et al., 2016; Mubarak et al., 2017) and offensive language (Wiegand et al., 2018; Zampieri et al., 2019). Prior studies have explored the use of aggressive and hateful language on different platforms such as Twitter (Xu et al., 2012; Burnap and Williams, 2015; Davidson et al., 2017; Wiegand et al., 2018), Wikipedia comments[1], and Facebook posts (Kumar et al., 2018a). Our present study is one of the first studies to make use of YouTube comments for computational modelling of aggression and misogyny (although there have been quite a few studies on pragmatic aspects of YouTube comments such as (Garcés-Conejos Blitvich, 2010; Garcés-Conejos Blitvich et al., 2013; Lorenzo-Dus et al., 2011; Bou-Franch et al., 2012)). Some of the earlier studies on computational modelling of misogyny have focussed almost exclusively on tweets ((Menczer et al., 2015; Frenda et al., 2019; Hewitt et al., 2016; Fersini et al., 2018b; Fersini et al., 2018a; Anzovino et al., 2018; Sharifirad and Matwin, 2019)). Also all of these studies have focussed on English or European languages like Italian and Spanish. And as such this is the first study on computational modelling of misogyny in two of India's largest languages - Hindi

---

[1]`http://bit.ly/2FhLMVz`

and Bangla.

In the following sections, we will discuss the corpus collection and annotation for this study and the development of a baseline misogyny classifier for the two languages.

## 2. The ComMA Project

The use of a wide range of aggressive and hateful content on social media becomes interesting as well as challenging to study in context to India which is a secular nation with religious as well as linguistic and cultural heterogeneity. The aim of the 'Communal and Misogynistic Aggression in Hindi-English-Bangla (ComMA) project' is to understand how communal and sexually threatening misogynistic content is linguistically and structurally constructed by the aggressors and harassers and how it is evaluated by the other participants in the discourse. We will use the methods of micro-level discourse analysis, which will be a combination of conversation analysis and the interactional model used for (im)politeness studies, in order to understand the construction and evaluation of aggression on social media.

We will use the insights from this study to develop a system that could automatically identify if some textual content is sexually threatening or communal on social media. The system will use multiple supervised text classification models that would be trained using a dataset annotated at 2 levels with labels pertaining to sexual and communal aggression as well as its evaluation by the other participants. The dataset will contain data in two of the largest spoken Indian languages - Hindi and Bangla – as well as code-mixed content in three languages – Hindi, Bangla and English. It will be collected from both social media (like Facebook and Twitter) as well as comments on blogs and news/opinion websites.

The research presented in this paper is carried out within the framework of this project and it focusses on one part of the project - automatic identification of misogyny.

## 3. Corpus Collection

### 3.1. Sources

For the purpose of the project, online sources laden with comments were carefully selected. In general, extensively used social media platforms were considered primary sources because of their massive footfall. Other than social media we also looked at some other popular streaming and sharing platforms. These were namely

- Facebook
- Twitter
- YouTube

The actual sources of information ranged from public posts, tweets, video blogs (vlogs), news coverage and so on. As a policy we did not invade into personal posts.

### 3.2. Sampling Criteria for Conversations

Given the desired output of the project and its requirements, conversations and opinions were selected on the basis of the points mentioned below

#### 3.2.1. The Volume of Conversation

In order to prepare a considerable dataset for training and looking at the requirement, only those posts and/or conversations were selected which were vastly commented by netizens. For example, in case of YouTube vlogs, news clips, opinions our team looked out for most commented among the chosen few at the initial stage. On an average, a minimum of 100-150 comments were kept as the benchmark for such cases.

#### 3.2.2. The Relevance Criterion

As we mentioned earlier, the choice of source materials was not random. Rather, a selection criterion was followed. After copious deliberations with the members, it was determined that we can only entertain those sources where misogyny is more likely to be expressed. A list of domains of possible source materials was considered and to name a few of those included the following -

- Women's fashion vlogs
- women's fitness tips videos
- news coverage of violent crime involving women
- celebrity news and gossip vlogs
- current socio-political commentaries and pertinent issues
- any other issue of immediate interest

India being a multilingual nation, it was observed that the content from any one source were in multiple languages. As such during the initial process of data collection from designated sources we needed to carefully separate content in different languages. Therefore, a language identification task was taken up with the native speakers and Hindi, Bangla, English were separated out and marked as such. A separate task was also carried out in order to separate Bangladeshi Standards and Indian standards of the Bangla data. Finally, the code-mixed English-Hindi and English-Bangla comments were separated out. The process of identification involved carefully analysed linguistically relevant information such as peculiar lexical choice, unique phonetic representation of chosen lexical items and regional colloquial usage.

This manual annotation of languages and varieties were used to develop an automatic language identification system for these languages. This system was developed using Support Vector Machines and uses word trigrams and character 5-grams for making the prediction about the language of the content. It achieved an F-score of 0.93 and has worked reasonably well for automatically classifying content into one of the languages before being sent to annotators or even misogyny and aggression classifiers.

# 4. The Aggression Tagset

In this section, we present the detailed guidelines for annotating the text from social media with information about aggression and misogyny. It gives a description of these categories and the features and, how those were employed during the annotation process. All annotations have been carried out at the level where the annotation target was a complete post, a comment or any one unit of the discourse. We would like to mention here that all of the data are represented as they were from the actual posts/sources. The authors and the project members do not bear ill feeling to people/names mentioned in the examples. Also, we do not endorse such aggressive and misogynistic language as one may find in the examples.

The aggression annotation was carried out using the aggression tagset (discussed in (Kumar et al., 2018b)). The tagset is reproduced in Table 1.

| TAG | AGGRESSION LEVEL |
|-----|------------------|
| OAG | Overtly Aggressive |
| CAG | Covertly Aggressive |
| NAG | Non - Aggressive |

Table 1: Aggression Annotation Tagset

# 5. The Misogyny Tagset

Misogyny identification is a binary classification task and the labels that we use for the task (Table 2) as well as the detailed guidelines (as developed and used by the annotators) are discussed below.

| TAG | ATTRIBUTE |
|-----|-----------|
| GEN | Gendered or Misogynous |
| NGEN | Non-gendered or Non-misogynous |

Table 2: Misogyny Annotation Tagset

## 5.1. Gendered or Misogynous (GEN)

This refers to such cases where verbal aggression aimed towards

- the stereotypical gender roles of the victim as well as the aggressor
- aggressive reference to one's sexuality and sexual orientation
- attacks the victim because of / by referring to her/his gender (includes homophobic and transgender attacks)
- includes attack against the victim owing to not fulfilling gender roles assigned to them or fulfilling the roles assigned to another gender

Some of the examples of this class are given below.

- tere ma se puch sale tera bap kon h [Go and ask your mother who your father is.]
- Napushank tha Nehru... lesbo thi indira [Nehru was impotent, Indira was a lesbian]
- Is hijray Rajnath ko chori pahna do [Give bangles to this eunuch Rajnath]

## 5.2. Non-gendered or Non-Misogynous (NGEN)

The text which is not gendered will be marked not gendered

## 5.3. Unclear (UNC)

In rare instances where it is not possible to decide whether the text is GEN or NGEN, this tag was be employed. It was not included in the final tagged document. It only served as an intermediary tag for flagging and resolving really ambiguous and unclear instances. For the sake of clarity and removing ambiguities in the annotation guidelines, an additional set of guidelines were formulated (as a result of discussion with the annotators). They are reproduced in the following sections -

## 5.4. General Instructions

The task relates to figuring out the 'intentionality' of the speaker (as manifested in the language used by her / him). You need to figure out if, something that is being said,

- arises out of an inherent bias of the speaker or
- an acceptance of that bias or
- propagates the bias (knowingly or unknowingly) or
- endorses the bias (again intentionally or unintentionally; or covertly or overtly)

The task could be approached by looking at the text and trying to figure out if it

- Attacks the victim because of / by referring to her/his gender (includes homophobic and transgender attacks) or
- includes attack against the victim owing to not fulfilling gender roles assigned to them or fulfilling the roles assigned to another gender

## 5.5. Attack against Women

Gendered does NOT mean any attack against women; it will be gendered only when the attack is BECAUSE of someone being a woman (or a man or a transgender or any of the countless gender identities). For example,

1. @KaDevender भडवा है साला खलिस्तानी और पाकिस्तानी एजेंट है और ये पाकिस्तानी स्लीपर सेल की मेंबर है यहाँ चुप के बैठी है किसी दिन बम बांध के कूद जाएगी और हजारो बेगुनाहो की जान ले लेगी ..निर्दोष हिन्दू के मारने पे ताली तो अभी बजाती है ..आप का चुसेन्द्र

He is a bastard, a Khalistani and Pakistani agent and she is a member of the Pakistani sleeper cell. She is hiding here and will jump with bomb anyday and kill thousands of innocent people...she appreciates the killing of innocent Hindus .. your sucker

2. Meye r maa eki character er chi

    daughter and mother are as same character, disgusting

In both (1) and (2), even though the attack is against a woman, the locus of attack may not be the gender. While in (2) the absence of a gender bias and misogyny is clear, in (1) it is little complicated because of the use of the last word and might be interpreted as gendered because of its use.

### 5.6. Jokes

One of the tests employed for resolving if a jone was gendered or not was to see if the gender of the target of the joke is changed, then the joke still works or not. If not, then it rests on some kind of gender bias. For example

1. Teacher : 'शीला कपडे पहन चुक' थी ' ईस वाक्य को अपनी भाषा में बोलो। Student -: भेनचोद लेट हो गए !!!!

    Teacher: 'Sheela had already worn her clothes' say this sentence in your own language. Student -: Fuck, we are late!!!!

It is 'Gendered' since the joke will not work for other gender

### 5.7. Satire/ sarcasm

A lot of times, for the lack of complete context, it was not clear if a comment was satire / sarcasm or not. Such unclear instances were initially tagged 'Unclear' and later a decision was arrived at through discussion among the annotators and, if required, based on voting. For example,

1. "Jithna dethe hein is d Benchmark for Jithna lena hai... - A Father (of Daughters and a Son) #Dowry #Jehaz #Shukrana #Nazrana #weddingideas #weddingseason #wedding #weddingdress #weddinggift"""

It is ambiguous. One of the ways of resolving such cases might be to look at hashtags and try to see the intention of the speaker. In this case, #Shukrana, #Nazrana etc seems to carry positive connotations. Also the tweet itself may look like a justification for dowry.

### 5.8. Poetry / shayari

If the intent was not clear in case of poetry then it was marked 'Unclear' and was later resolved using a majority voting. However, in other instances, it was marked as perceived by the annotators For example,

1. सिमटकर चूड़ियों में छुपने लगा शायद मैंने जो चूम लिया उसको मुझको चुभने लगा शायद मैंने जब आगोश में भर लिया उसको मुझसे जलता है तेरा कंगन शायद... मैंने जो इश्क कर लिया है उसको #साहिब१ #चूड़ी #कंगन #हिंदी_शब्द #शब्दनिधि

    She hid herself in her bangles probably, when I kissed her, it started to prick me when I hugged her, your bangle is envious of me probably because I made love to you.

2. Ranuu goo Ranuu Lagboo tmr Karr Nunuu?? Himeshh Salmann nakii Sonuu??

    "oh Ranu! who's penis do you want? Himesh Salman or Sonnu??

In (1), the poetic verse is romantic in nature and talk about lovemaking. Such expressions can be gendered and express misogyny if they clearly represent lack of consent. Because the axis of consent is not clear here we do not mark it as Gendered. (2), however, is clearly gendered, despite being in verse (not really poetry, though) since the imagery of sex and sexual violence is unnecessarily invoked for attacking the victim.

### 5.9. Figuring out tacit intentions / underlying bias

In some cases, at the surface, speakers may seem to be speaking against a biased practice / behaviour but the arguments given by her / him may not actually be questioning those biases itself and might even be creating another kind of bias. Let us take a look at the following examples,

1. @ aajtak @ News18India @ sdtiwari Time has come that a debate on #Dowry should be organised on highest level. it is absolutely essential to abolish #Dowry from Hindu Society. A honest hard worker can't manage to satisfy Groom's demand, particularly when #Bride is highly educated.

2. Don't Support #Dowry at all.Thre is no point to strt a rltnshp on exchnge of Bt also nd to tch society ,all failed marriages r nt due to #Dowry.So stop nmng every broken marriage as #FakeCases_498A_DV_125_377_376 Real sufferers nvr gts justice,help them stop misuse of #laws

3. So according to you protesting against molestation is a crime ? Sir Don't you have any daughters or sister? #BHU_लाठीचार्ज #bhu_molestation

4. When a thousand years old #Hindu tradition is followed in #Kerela then Muslims came forward to say that it oppresses women's freedom, even Hindu Women" themselves says that they don't want to enter #Sabrimala & respect the traditions! #IslamExposed https:// twitter.com/theskindoctor1 3/status/1113435724269981696 ..."

5. भेनचोद ये गुलाबी पैंट कौन पहनता है बे

   Fuck man, who wears a pink trouser?

6. हम देश वासी जवानों के जित का #Abhinandan करते हैं। अब हम सबको मिलकर #SpecialStarus4Jawan सुनिश्चित करना होगा। जो अपने जान जोखिम में डालकर देश की रक्षा कर रहा, खुद को देश के लिए समर्पित कर दिया है, उसके लिए यह तो होना ही चाहिए। जवानों को #Dowry Act से बाहर करो @ ani @ dna @ aajtakpic.twitter.com/ezmfDEzxXQ

   We the people of this nation #Abhinandan (welcome) the victory of our soldiers. Now we all should ensure a#SpecialStarus4Jawan. The one who is protecting our country by endangering their lives, has donated his life for the cause of the nation, this should definitely be done for him. Exempt soldiers from #Dowry Act.

In tweet 1 and 2, the dissatisfaction is because of the inability to afford the demands (and not because the 'demand' itself is discriminatory and biased). So its a financial argument for an inherently 'gender' issue since only women are supposed to give dowry. Also it creates a distinction between 'educated' and 'uneducated' girls, thereby, implying that dowry is fine for undeducated girls. Thus it creates another bias (which clearly doesnt exist for the other gender). Thus even though it seems to be opposing a gendered practice like dowry; it doesnt actually oppose the underlying bias in a practice like this. While (3) this looks like a support for protest against molestation, it reinforces the stereotype of women as sisters and daughters. Also molestation is a crime and it doesnt have to do anything with whether there are other women in someone's life or not. On the face of it, (4) may look like a religious comment. However, underlyingly an attempt is made here to present a gender issue as a religious issue. It is a support for a practice which is biased against a specific gender (and religion is used as a smokescreen for propagating that bias). (5) reinforces the stereotypical gender associated with the use of a particular colour by a particular gender. (6) Puts gender issues vis-a-vis army which is not at all relevant or comparable and favours a certain kind of preferential treatment based on job. Its a support for dowry in certain cases (since dowry is not considered a gendered act by the speaker).

### 5.10. Abuses

In general, abuses involving sex and sexual organs will be considered gendered since they emanate from an inherent gender bias. Let us take a look at the following examples -

1. @USER This game sucks donkey balls

2. bitch calm down you pussy when yo ppl ain't around

3. अबे ओ अपनी बहन से पैदा कीड़े । भड़ुआ बनना है तो पप्पू के लुंड पर बैठ।तेरी अम्मी का यार मत समझ मुझे झोपडी के। संघी आतंकवादी होते है क्या।अपनी बहन का हला ला करने कहीं और जा ।तेरे जैसे 10 रोज ठिकाने लगाता हूं। समझा नपुंसक

   Hey you, a worm born out of your sister. If you wish to be a fucker then go sit on Pappu's penis. Do not think of me as your mother's boyfriend. Those who belong to the Sangh are not terrorists. Go somewhere else to perform your sister's halala. I deal with the likes of you everyday. Do you understand you impotent.

4. बॉसडीके, मधरचूत, तेरी माँ की, बहिन की छूट, रंडी का ौलत, खानदानी रंडी का ौलत, हीरामंडी का पिल्ला, भादवा लौड़ा लुंड कमीना, छूट के ढक्कन, छिपकली के गांड के पसीने

   Motherfucker, your mother's your sister's pussy, son of a bitch, litter of heeramandi, pussy cap, sweat of the anus of a lizard.

5. চশমাপড়া মাসীমার গুদের নাম্বার টা কি জানতে পেরেছিলেন ?

   Did you get the number of bespectacled aunty's vagina?

6. भेनचोद ये गुलाबी पैंट कौन पहनता है बे

   Fuck man, who wears a pink trouser.

Even though there is no direct attack in (1), the abuse here arises out of an understanding about what is considered an homosexual act. The abuses used in (3) show the biased and misogynistic outlook of the speaker. Even though the attack is not because of the gender, it carries the connotations of attack against a specific gender as it reinforces the role of women as sexual objects. At the same time it propagates the stereotypical ideas of honor, masculinity, etc. Abuses like those in (4) and (5) evoke sexual imagery and are used for attacking someone, hence, gendered. In (6) the abuse is just an exclamation marker and so not directed towards anyone. As such it is not gendered because of the use of this abuse (but see above for description of what makes it gendered).

### 5.11. Victim blaming

In a lot of cases of discussion around gender, it is the girls or the girls' side that are attacked - it is important to figure out the cases of blaming the victim for the problems they are facing (because of the patriarchal societal structure). For example,

1. #DAHEZ LDKIYO K MAA BAAP HI DETE HAI, JB KOI V CHEEZ AISE HI MIL JAAYE TO LOG Q NAA LE. AB MERE SAATH HI HAI MAI JISSE PYAR KRTA HU USKI SAADI 1 GOVT. JOB WAALE SE HO RHI H AND THEY ARE TAKING #DOWRY. BUT I AM AGAINST DOWRY, I JUST WANT HER ONLY. But govt. Job is in b/w

   Dowry is gifted by the bride's parents only. When something is received without a price then why should'nt one take it? Now look at my case. The one I love is getting married to a government employee and they are taking #DOWRY. BUT I AM AGAINST DOWRY, I JUST WANT HER ONLY. But govt. Job is in b/w

In this tweet, the speaker asserts that he is against dowry. However he still blames the parent of the girls for this kind of practice and at the same time also absolves the boys of any responsibility. Such cases of victim blaming is gendered.

### 5.12. Description of an event / fact

Describing a gendered act / incident / practice does not make the text gendered. In such cases, it will be gendered only if the speaker endorses the action or depicts an underlying bias. Let us take a look at the following examples -

1. IF YOU SAY ONLY #MOTHERINLAW #HUSBAND ARE ACCUSED. YOU ARE TOTALLY WRONG. BECAUSE IT'S HER OWN PARENT WHO MARRIED TO THAT GUY AND FORSURE EARLIER HER FATHER HAD GIVEN #DOWRY TO THEM. SO, HER PARENTS ARE ALSO INVOLVED. EVERY PARENTS WANT GOVT. JOB GUY AND PAY. AGAINST THIS SYSTEM.

2. Against the grain: In some parts of #Maharashtra, women get #dowry https://trib.al/gz1NTix

3. If the groom's family in China is unable to afford the bride prices, then he is not considered a good match. Learn more: https://buff.ly/2CUDzqv #China #marriage-market #matchmaking #dowry #brideprices #culturepic.twitter.com/v8MxjGsQz2

4. People were often coupled in European countries according to class and, thus, economic advantage. Learn more here: https://buff.ly/2umI6Nu #economicadvantage #dowry #Europe #culture #marriagepic.twitter.com/Pas1rKavLk

5. जब बर्तन मांज कर आयी वो तो गालों ने बताया..!! कि बर्तन काँच का कोई आज फिर से टूटा गया..!! @YadavsAniruddh @Anjupra7743 @KaranwalTanu @AmbedkarManorma follow @Rana11639322

   When she came after cleaning dishes her cheeks revealed it all..!! that a glass dish has has been broken again today..!!

(1) describes a biased situation. However there is no evidence to show that the speaker also endorses it. As such even though the situation being described is gendered, the tweet itself is not. (2) doesnt question the gender bias in the dowry system and acts as an underlying support for dowry. Irrespective of who pays the dowry to whom – its always biased against a specific gender. Since the speaker seems to be endorsing this view, it is gendered. In (3) even though it may look like the description of a practice, the underlying intention of the speaker is to support and justify the practice of dowry by giving a parallel example from a different context. (4) is presented as a covert support for the dowry system, which puts one specific gender in a very disadvantageous position and as such the tweet itself is gendered as well. In (5) even though the incident being described is gendered, the tweet is not a support for that. Thus it will not be gendered.

### 5.13. Mixed bias

In some cases, gender bias might be mixed with other kinds of biases (like religious or regional). In such case, if gender bias is also present, it will be marked gendered. For example,

1. Arnab @republic is visibly anti ChristoROPcommieFascists. But the #MeToo / Libtard women hv wrapped him in their fingers. So in their appeasement he took anti Hindu stand on #Sabrimala . Appeased LGBTQ during Section 377. Vilified the accused in #MeToo b4 Court Trial.

In this case, religion seems to be the locus of attack. However, it attacks a lot of other instances of support for non-male rights, hence, biased for a specific gender.

### 5.14. Other Ambiguities

Let us take a look at the following examples -

1. http://chng.it/DPFHRS9B4T.Please … sign this petition. For men and their families falsely accused in #DomesticViolence, #dowry and #498a by leeching women, there are no laws to give them a fair trial and no laws to punish leeching women. #MenCommission and #GenderEquality in laws needed.

2. Next surgical strike she along with her entire terrorist clan shd be dropped in #Napakistan #Disgusting she is. She also orchestrated fake #Asifa narrative. Shameless ppl dance on dead bodies..

3. Should we go for GENDER INJUSTICE here? #sabrimala was the same But as I respect my religion and its beliefs i fully support this ritual and i am fully satisfied with whatever rule is imposed. Jay matadi

(1) is a call to punish those who misuse the law and so apparently promoting gender equality. However when it is accompanied by a call to form a men commission, it seems to be ignoring and undermining the issues that a woman faces. There are several laws that are misused by several people - however it is the one law that is intended to protect the women that causes the maximum uproar. However, having said this, the intention of the speaker does not seem to be biased. In such cases, the annotators may annotate based on their intuition on case-by-case basis or mark it as 'unclear' so that annotations by multiple annotators may be taken. In such cases, they must also include a comment describing the ambiguity. In (2), the question to settle is this – is the criticism BECAUSE the person being criticised is a man / woman or the criticism is directed somewhere else? In this case, the criticism doesnt seem to be directed at gender. However bringing in #Asifa and calling it fake shows a gender bias. Such cases also have to be handled as mentioned above. In (3), the stand taken by the speaker is not clear here and as such may be marked unclear

## 6. Annotation of the Dataset

While developing a tagset is a relatively easy and a regular task for such automatic detection, assigning tags to units is not. Annotation of data involves multiple human intervention and hence, the need of constant deliberations over the justification of assigned tags is a much required step. The issues that we face in annotation occur due to different level of understanding of the language in question or personal prejudices and bias over interpretation and so on. Basically, it involves the differing worldview of many individuals.

Notwithstanding the personal interpretations, there were occasions where reaching to a consensus was hard in this task. As the task involved more than one individual, the device of sample annotation process was utilised and all the members were given a single file to tag. Although, in about 75 per cent or more cases the tags were unanimous, some data required special attention as different individuals tagged these cases differently. In such cases a three-way process was developed in the course of deliberations. It is as follows, 1) Counterexample method is use to test the comment: 2) Annotators vote are looked at: All the collaborators joined in conference to deliberate over the data in question. Independent members were also consulted in the process to get a different view. Native speakers took part to disambiguate examples or provide explanations for parts not understood. Finally, a vote on the most relevant interpretation was carried on to reach to a consensus 3) Less strict tag is assigned: Instead of marking such questionable data with absolute tags (as are in the tagset), at times a less stringent approach was taken up. In this the annotators were asked to mark such data either as DOUBTFUL or keep them untagged for a discussion later. This helped immensely in the smooth and timely flow of the annotation process as well as later resolution through discussion. 4) As was mentioned in the corpus collection section, the team members often encountered Bangla data both from Bangladeshi and Indian netizens. In such cases a separate tagging criterion was assigned for BNGB (Bangladeshi Bangla) and BNG (Indian Bangla). The process of separating the two varieties followed regulation linguistic practices in analysis by pointing out structural and usage based uniqueness of the two regions. In our project we have utilised the Indian Bangla data while keeping the Bangladeshi variety aside for the moment.

## 7. The Final Dataset

The final dataset contains a total of over 25,000 comments in the 3 languages - Hindi, Bangla and English. Figure 1 shows the share of data in each language. Overall, almost 3,000 (over 11%) are gendered / misogynistic and more than 23,000 are non-gendered. The proportion of gendered comments in Hindi, Indian Bangla and Hindi-English code-mixed comments hovers around 10 - 15%, while in English it is just around 4%. A language-wise break-up and comparison is given in Figure 2.

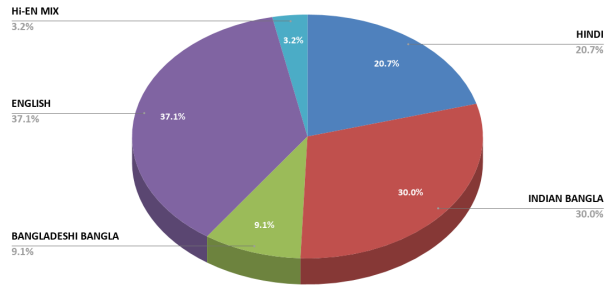

Figure 1: Languages in the Dataset

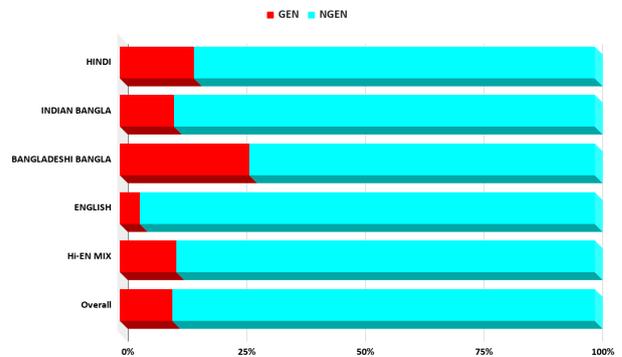

Figure 2: Misogyny in the Dataset

Almost half of these comments in Hindi, Indian Bangla and English are also annotated for 3 levels of aggression. A language-wise break-up and comparison of aggressive comments in the dataset is given in Figure 3 The share of aggressive (taking together both overtly and covertly aggressive comments) comments in the dataset is around 45% of the total annotated dataset in Hindi and Indian Bangla, while it is around 20% in English. These are similar to what was reported in (Kumar et al., 2018b).

We also took a look at the co-occurrence of aggressive and gendered comments to see if most of the gendered / misogynous comments are also generally aggressive or not. Overall, it turns out that over 80% of the gendered comments are also aggressive; on the other hand, less than 30% of non-gendered comments are aggressive. It shows that misogyny may be strongly correlated with aggression and even though a substantial proportion of non-gendered comments are also aggressive (in our dataset), a much larger proportion of gendered comments are aggressive. A language-wise break-up of proportion of aggression in gendered as well as non-gendered comments are given in Figure 4 and Figure 5.

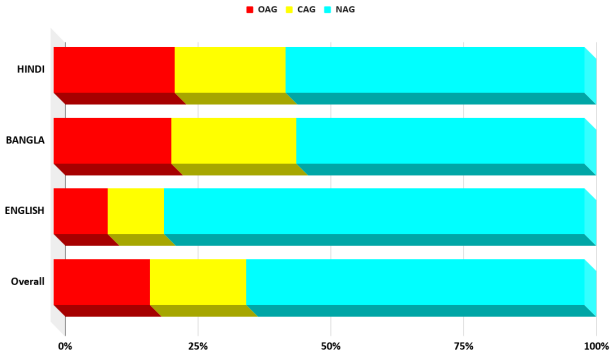

Figure 3: Aggression in the Dataset

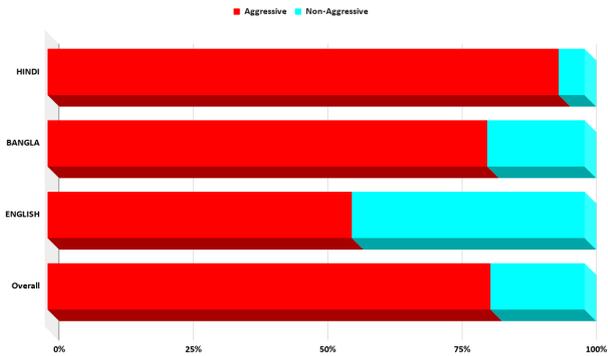

Figure 4: Co-occurrence of Misogyny and Aggression

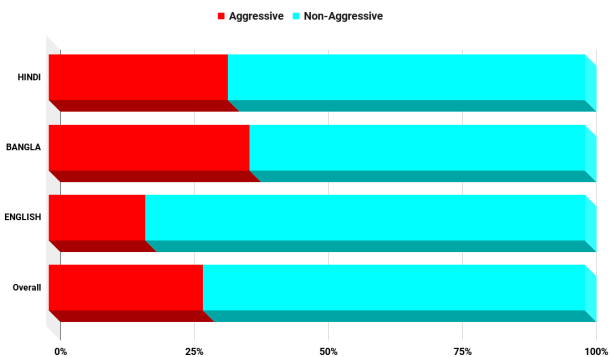

Figure 5: Co-occurrence of Non-gendered and Aggression

## 8. Baseline Misogyny Classifier

Using a subset of the annotated dataset, we trained Support Vector Machine (SVM) for automatic identification of misogyny in Hindi, Bangla and English (in the Indian context). The statistics of dataset used for training and testing is given in Table 3. We experimented with different combinations of word (uni, bi and tri) and character (2 - 5) n-grams as features. We carried out a 10-fold cross validation and also experimented with the C-value of SVM ranging from 0.001 to 10. The best performing classifiers, along with their performance for each of the three languages is summarised in Table 4.

| LANGUAGE | GEN | NGEN | TOTAL |
|---|---|---|---|
| Hindi | 828 | 3,156 | 3,984 |
| Bangla | 871 | 2,955 | 3,826 |
| English | 393 | 3,870 | 4,263 |

Table 3: Training and Testing Dataset

| Language | Character n-gram | Word n-gram | F-Score |
|---|---|---|---|
| Hindi | 3 | 3 | 0.87 |
| Bangla | 5 | NA | 0.89 |
| English | 2 | NA | 0.93 |

Table 4: Baseline Classifier Result

As is evident from this, character and word n-grams prove to be quite a string baseline, which achieves an f-score close to 0.90 for Hindi and Bangla and for English it achieves an impressive score of 0.93.

## 9. Summing Up and the Way Ahead

In this paper, we have discussed the development of a multilingual corpora in Hindi, Bangla and English, annotated with the information about it being gendered or not. The total corpus consists of more than 25,000 comments from different YouTube videos annotated with this information. The dataset has been made publicly available for research purposes [2]. We also trained a baseline classifier on this dataset which gives a high f-score of over 0.87 for Hindi, 0.89 for Bangla and 0.93 for English dataset.

We are currently working on expanding the dataset to include data from other platforms and domains and then test the classifier to see how well it performs across different kinds of dataset. Our goal is to have a dataset of at least 50,000 comments / units in each of the three languages and develop a multilingual classifier that could work reasonably well for different platforms / domains in automatically detecting misogny over social media.

---

[2] The dataset has been publicly released via a shared task on aggression and misogyny identification - https://sites.google.com/view/trac2/shared-task

## 10. Acknowledgements

We would like to thank Facebook Research for an unrestricted research gift for carrying out this research.

## 11. Bibliographical References

Agarwal, S. and Sureka, A. (2015). Using knn and svm based one-class classifier for detecting online radicalization on twitter. In *International Conference on Distributed Computing and Internet Technology*, pages 431 – 442. Springer.

Agarwal, S. and Sureka, A. (2017). Characterizing linguistic attributes for automatic classification of intent based racist/radicalized posts on tumblr microblogging website.

Anzovino, M., Fersini, E., and Rosso, P. (2018). Automatic identification and classification of misogynistic language on twitter. In Max Silberztein, et al., editors, *Natural Language Processing and Information Systems*, pages 57–64, Cham. Springer International Publishing.

Bou-Franch, P., Lorenzo-Dus, N., and Blitvich, P. G.-C. (2012). Social Interaction in YouTube Text-Based Polylogues: A Study of Coherence. *Journal of Computer-Mediated Communication*, 17(4):501–521, 07.

Burnap, P. and Williams, M. L. (2015). Cyber hate speech on twitter: An application of machine classification and statistical modeling for policy and decision making. *Policy & Internet*, 7(2):223–242.

Cambria, E., Chandra, P., Sharma, A., and Hussain, A. (2010). Do not feel the trolls. In *ISWC, Shanghai*.

Dadvar, M., Trieschnigg, D., Ordelman, R., and de Jong, F. (2013). Improving Cyberbullying Detection with User Context. In *Advances in Information Retrieval*, pages 693–696. Springer.

Davidson, T., Warmsley, D., Macy, M., and Weber, I. (2017). Automated Hate Speech Detection and the Problem of Offensive Language. In *Proceedings of ICWSM*.

Djuric, N., Zhou, J., Morris, R., Grbovic, M., Radosavljevic, V., and Bhamidipati, N. (2015). Hate Speech Detection with Comment Embeddings. In *Proceedings of WWW*.

Fersini, E., Nozza, D., and Rosso, P. (2018a). Overview of the evalita 2018 task on automatic misogyny identification (AMI). In Tommaso Caselli, et al., editors, *Proceedings of the Sixth Evaluation Campaign of Natural Language Processing and Speech Tools for Italian. Final Workshop (EVALITA 2018) co-located with the Fifth Italian Conference on Computational Linguistics (CLiC-it 2018), Turin, Italy, December 12-13, 2018*, volume 2263 of *CEUR Workshop Proceedings*. CEUR-WS.org.

Fersini, E., Rosso, P., and Anzovino, M. (2018b). Overview of the task on automatic misogyny identification at ibereval 2018. In Paolo Rosso, et al., editors, *Proceedings of the Third Workshop on Evaluation of Human Language Technologies for Iberian Languages (IberEval 2018) co-located with 34th Conference of the Spanish Society for Natural Language Processing (SEPLN 2018), Sevilla, Spain, September 18th, 2018*, volume 2150 of *CEUR Workshop Proceedings*, pages 214–228. CEUR-WS.org.

Frenda, S., Ghanem, B., Montes-y Gómez, M., and Rosso, P. (2019). Online hate speech against women: Automatic identification of misogyny and sexism on twitter. *Journal of Intelligent & Fuzzy Systems*, 36(5):4743–4752.

Garcés-Conejos Blitvich, P., Lorenzo-Dus, N., and Bou-Franch, P. (2013). Relational work in anonymous, asynchronous communication: A study of (dis)affiliation in youtube. In Istvan Kecskes et al., editors, *Research Trends in Intercultural Pragmatics*, pages 343–366. De Gruyter Mouton, Berlin.

Garcés-Conejos Blitvich, P. (2010). The youtubification of politics, impoliteness and polarization. In Rotimi Taiwo, editor, *Handbook of Research on Discourse Behavior and Digital Communication: Language Structures and Social Interaction*, pages 540 – 563. IGI Global, USA.

Greevy, E. and Smeaton, A. F. (2004). Classifying racist texts using a support vector machine. In *Proceedings of the 27th annual international ACM SIGIR conference on Research and development in information retrieval*, pages 468 – 469. ACM.

Greevy, E. (2004). *Automatic text categorisation of racist webpages*. Ph.D. thesis, Dublin City University.

Hewitt, S., Tiropanis, T., and Bokhove, C. (2016). The problem of identifying misogynist language on twitter (and other online social spaces). In *Proceedings of the 8th ACM Conference on Web Science*, WebSci '16, page 333–335, New York, NY, USA. Association for Computing Machinery.

Kumar, S., Spezzano, F., and Subrahmanian, V. (2014). Accurately detecting trolls in slashdot zoo via decluttering. In *Proceedings of IEEE/ACM International Conference on Advances in Social Networks Analysis and Mining (ASONAM)*, pages 188–195.

Kumar, R., Ojha, A. K., Malmasi, S., and Zampieri, M. (2018a). Benchmarking Aggression Identification in Social Media. In *Proceedings of TRAC*.

Kumar, R., Reganti, A. N., Bhatia, A., and Maheshwari, T. (2018b). Aggression-annotated corpus of hindi-english code-mixed data. In Nicoletta Calzolari (Conference chair), et al., editors, *Proceedings of the Eleventh International Conference on Language Resources and Evaluation (LREC 2018)*, Paris, France, may. European Language Resources Association (ELRA).

Kwok, I. and Wang, Y. (2013). Locate the Hate: Detecting Tweets Against Blacks. In *Proceedings of AAAI*.

Lorenzo-Dus, N., Blitvich, P. G.-C., and Bou-Franch, P. (2011). On-line polylogues and impoliteness: The case of postings sent in response to the obama reggaeton youtube video. *Journal of Pragmatics*, 43(10):2578 – 2593. Women, Power and the Media.

Malmasi, S. and Zampieri, M. (2017). Detecting Hate Speech in Social Media. In *Proceedings of the International Conference Recent Advances in Natural Language Processing (RANLP)*, pages 467–472.

Malmasi, S. and Zampieri, M. (2018). Challenges in discriminating profanity from hate speech. *Journal of Experimental & Theoretical Artificial Intelligence*, 30:1 – 16.

Menczer, F., Fulper, R., Ciampaglia, G. L., Ferrara, E., Ahn, Y., Flammini, A., Lewis, B., and Rowe, K. (2015). Misogynistic Language on Twitter and Sexual Violence. In *Proceedings of the ACM Web Science Workshop on Computational Approaches to Social Modeling (ChASM)*. Association of Computing Machinery, 1.

Mihaylov, T., Georgiev, G. D., Ontotext, A., and Nakov, P. (2015). Finding opinion manipulation trolls in news community forums. In *Proceedings of the Nineteenth Conference on Computational Natural Language Learning, CoNLL*, pages 310–314.

Mojica de la Vega, L. G. and Ng, V. (2018). Modeling trolling in social media conversations. In *Proceedings of the Eleventh International Conference on Language Resources and Evaluation (LREC 2018)*, Miyazaki, Japan, May. European Language Resources Association (ELRA).

Mubarak, H., Kareem, D., and Walid, M. (2017). Abusive language detection on Arabic social media. In *Proceedings of ALW*.

Nitin, Bansal, A., Sharma, S. M., Kumar, K., Aggarwal, A., Goyal, S., Choudhary, K., Chawla, K., Jain, K., and Bhasinar, M. (2012). Classification of flames in computer mediated communications. *International Journal of Computer Applications*, 14(6).

Nobata, C., Tetreault, J., Thomas, A., Mehdad, Y., and Chang, Y. (2016). Abusive Language Detection in Online User Content. In *Proceedings of the 25th International Conference on World Wide Web*, pages 145–153. International World Wide Web Conferences Steering Committee.

Sax, S. (2016). Flame Wars: Automatic Insult Detection. Technical report, Stanford University.

Sharifirad, S. and Matwin, S. (2019). When a tweet is actually sexist. A more comprehensive classification of different online harassment categories and the challenges in NLP. *CoRR*, abs/1902.10584.

Waseem, Z., Davidson, T., Warmsley, D., and Weber, I. (2017). Understanding Abuse: A Typology of Abusive Language Detection Subtasks. *Proceedings of ALW*.

Wiegand, M., Siegel, M., and Ruppenhofer, J. (2018). Overview of the GermEval 2018 Shared Task on the Identification of Offensive Language. In *Proceedings of GermEval*.

Xu, J.-M., Jun, K.-S., Zhu, X., and Bellmore, A. (2012). Learning from Bullying Traces in Social Media. In *Proceedings of NAACL*.

Zampieri, M., Malmasi, S., Nakov, P., Rosenthal, S., Farra, N., and Kumar, R. (2019). Predicting the type and target of offensive posts in social media. In *Proceedings of the Annual Conference of the North American Chapter of the Association for Computational Linguistics: Human Language Technology (NAACL-HLT)*.